\documentclass[10pt,conference,a4paper]{IEEEtran}
\def\year{2018}\relax
\usepackage{epsfig}
\usepackage[T1]{fontenc}
\usepackage{color}
\usepackage[utf8]{inputenc}
\usepackage{amssymb}
\usepackage{amsmath}
\usepackage{bm}
\usepackage{dsfont}
\usepackage{algorithm}
\usepackage{amsthm}
\usepackage{graphicx}
\usepackage{algpseudocode}
\usepackage{caption}
\usepackage{subcaption}

\IEEEoverridecommandlockouts

\newtheorem{thm}{Theorem}
\newtheorem{defn}{Definition}

\title{Dynamic Ensemble Active Learning: \\A Non-Stationary Bandit with Expert Advice}

\author{\IEEEauthorblockN{Kunkun Pang\IEEEauthorrefmark{1}, Mingzhi Dong\IEEEauthorrefmark{2}, Yang Wu\IEEEauthorrefmark{3}\thanks{This work was supported by JSPS KAKENHI Grant Number 15K16024. It has been accepted at ICPR 2018 and won Piero Zamperoni Best Student Paper Award. The corresponding author is Yang Wu.}, Timothy M.  Hospedales\IEEEauthorrefmark{1}}
\IEEEauthorblockA{\IEEEauthorrefmark{1}University of Edinburgh, \ \ Email: k.pang@ed.ac.uk, t.hospedales@ed.ac.uk}
\IEEEauthorblockA{\IEEEauthorrefmark{2}University College London, \ \ Email: mingzhi.dong.13@ucl.ac.uk}
\IEEEauthorblockA{\IEEEauthorrefmark{3}Nara Institute of Science and Technology, \ \ Email: yangwu@rsc.naist.jp}
}

\begin{document}

\maketitle


\begin{abstract}
Active learning aims to reduce annotation cost by predicting which samples are useful for a human teacher to label. However it has become clear there is no best active learning algorithm. Inspired by various philosophies about what constitutes a good criteria, different algorithms perform well on different datasets. This has motivated research into ensembles of active learners that learn what constitutes a good criteria in a given scenario, typically via multi-armed bandit algorithms. Though algorithm ensembles can lead to better results, they overlook the fact that not only does algorithm efficacy vary across datasets, but also during a single active learning session. That is, the best criteria is non-stationary. This breaks existing algorithms' guarantees and hampers their performance in practice. In this paper, we propose dynamic ensemble active learning as a more general and promising research direction. We develop a  dynamic ensemble active learner based on a non-stationary multi-armed bandit with expert advice algorithm. Our dynamic ensemble selects the right criteria at each step of active learning. It has theoretical guarantees, and shows encouraging results on $13$ popular datasets.
\end{abstract}

\section{Introduction}

The key barrier to scaling or applying supervised learning in practice is often the cost of obtaining sufficient annotation. Active Learning (AL) aims to address this by designing query algorithms that effectively predict which points will be useful to annotate, thus enabling efficient allocation of human annotation effort. There are many different AL algorithms, each with appealing -- yet completely different -- motivations for what constitutes a good question to ask underpinning their design. For example, uncertainty or margin-based sampling \cite{Lewis:1994:SAT:188490.188495,Tong:2002:SVM:944790.944793} suggests querying the most uncertain or ambiguous point that is the closest point to the decision boundary. Expected error reduction  \cite{Roy:2001:TOA:645530.655646,hospedales2012dpea} queries points that the current model predicts will reduce its future error. Another typical approach is to label the most representative samples \cite{journals/jair/CohnGJ96,conf/kdd/2012,Yu:2006:ALV:1143844.1143980} to ensure the major clusters within the dataset are correctly estimated. {Besides these approaches, query-by-committee active learning queries points based on the disagreement between a committee of classifiers \cite{Seung:1992:QC:130385.130417,Abe:1998:QLS:645527.657478,loy2012stream}. More recent studies investigated hybrid criteria that balance multiple motivations \cite{conf/nips/2010,Wang:2015:QDR:2737800.2700408, DBLP:journals/tgrs/WangDZZJ17}.}

{These are all good ideas, yet there are situations where each is ineffective. For example, if the classes are heavily overlapped in an area of feature-space, uncertainty sampling will be tied up querying points in an impossible to solve region. If the current model is very poor, expected error reduction cannot accurately estimate its own future error. If the main data clusters are already well classified, representative sampling focused approaches may not fine-tune between them.}

These thought experiments are reflected empirically. The best algorithm for pool-based AL in practice varies both across datasets and also with the progress of learning within a given dataset \cite{Baram:2004:OCA:1005332.1005342,hsu2015active}. This observation has motivated research into both learning dataset and time-specific weightings for an AL algorithm ensemble. \cite{Donmez2007,journals/tkde/HospedalesGX13} developed heuristics for switching AL algorithms that are typically good at early versus late stage learning. In contrast,  \cite{Baram:2004:OCA:1005332.1005342,hsu2015active} 
developed methods for rapid online learning of a dataset-specific weighting for algorithms within an AL-ensemble.

The key insight of the Combination of Active Learning Online (COMB) \cite{Baram:2004:OCA:1005332.1005342} and Active Learning by Learning (ALBL) \cite{hsu2015active} algorithms is to formalise the query criteria selection task as a multi-armed bandit (MAB) problem. MAB problems have been well studied and many powerful algorithms with optimality guarantees exist. For example, if each query criterion in the ensemble is considered to be a bandit arm, and the learning improvement after executing a criterion is considered to be the arm's reward, then MAB algorithms such as EXP3 (Exponential-weight algorithm for Exploration and Exploitation) \cite{auer2002nonstochastic} can be applied to quickly learn the efficacy of  the arms (AL criteria) and is guaranteed to achieve a near optimal overall reward (learning improvement). A variant of this is to consider data-points to be arms, and AL criteria to be experts providing advice about promising arms. Then MAB with \emph{expert advice} algorithms such as EXP4.P (Exponential-weight algorithm for Exploration and Exploitation using Expert advice with high probability regret bound) \cite{beygelzimer2011contextual} optimise exploration and exploitation of experts, and achieve provably near optimal reward.

\begin{figure}[t]
\captionsetup[subfigure]{justification=centering}
    \centering
    \begin{subfigure}{0.5\columnwidth}
		\includegraphics[width=0.98\linewidth]{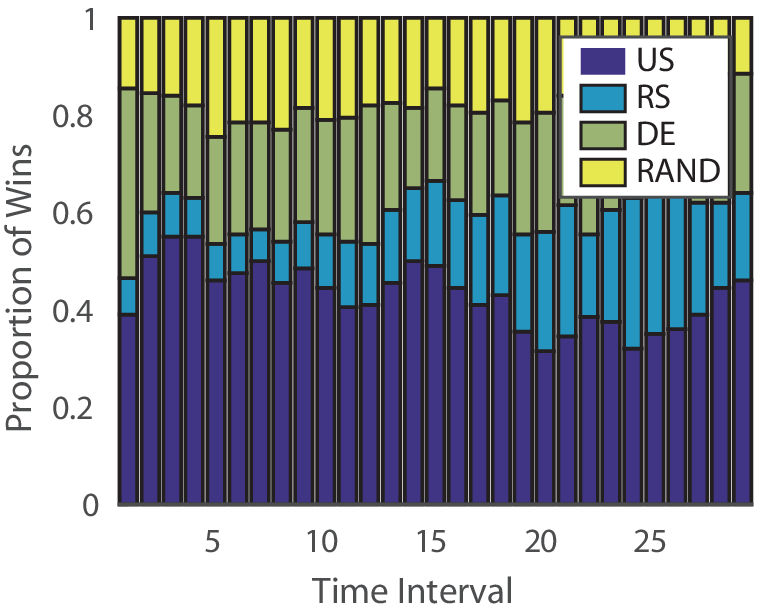}
		\caption{Proportion of Wins:\newline ``fourclass''}
		\label{fig1:UCI ND a}
	\end{subfigure}%
	\begin{subfigure}{0.5\columnwidth}
		\includegraphics[width=0.98\linewidth]{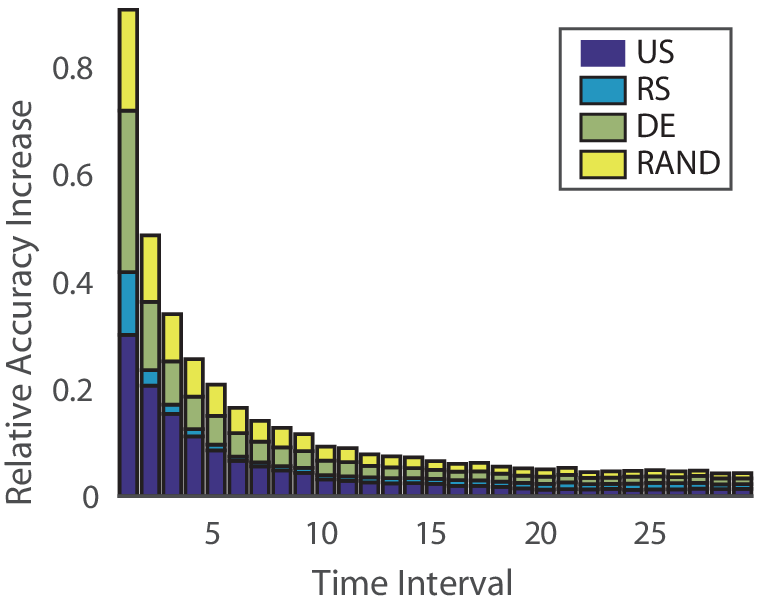}
		\caption{Relative Accuracy \newline Increments: ``fourclass''}
		\label{fig1:UCI ND b}
	\end{subfigure}
	\begin{subfigure}{0.5\columnwidth}
        \includegraphics[width=0.98\linewidth]{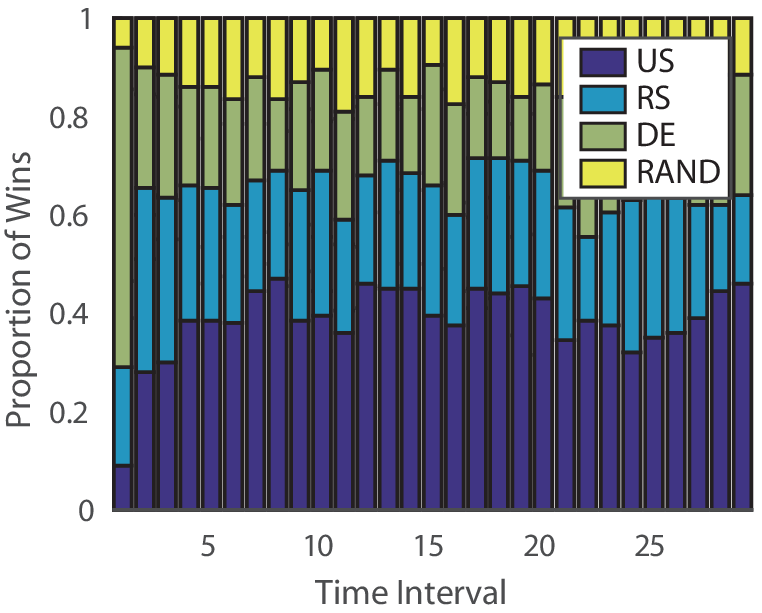}
        \caption{Proportion of Wins:\newline ``ILPD''}
		\label{fig1:UCI ND c}
	\end{subfigure}%
	\begin{subfigure}{0.5\columnwidth}
		\includegraphics[width=0.98\linewidth]{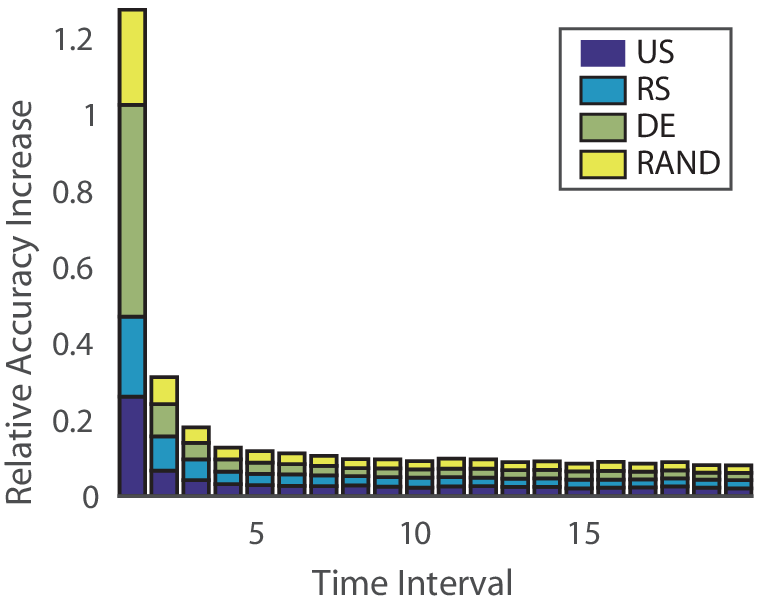}
		\caption{Relative Accuracy \newline Increments: ``ILPD''}
		\label{fig1:UCI ND d}
	\end{subfigure}
	\begin{subfigure}{0.5\columnwidth}
		\includegraphics[width=0.98\linewidth]{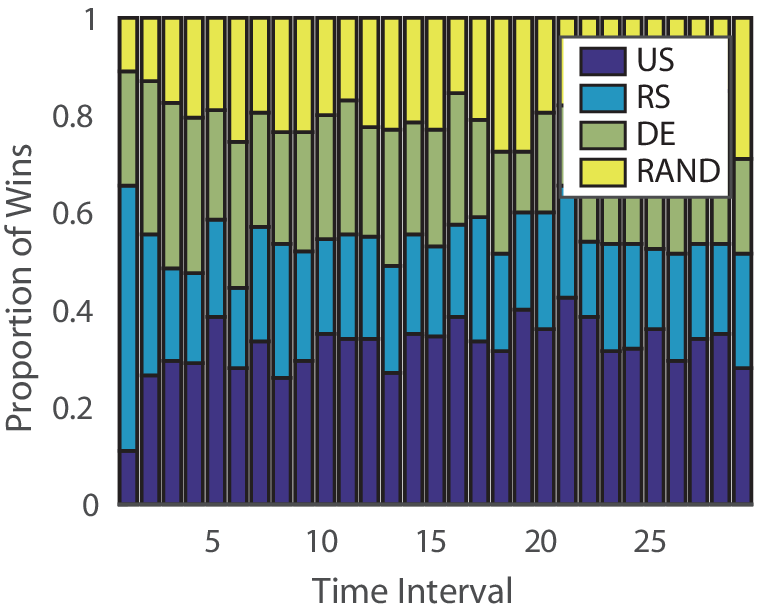}
		\caption{Proportion of Wins:\newline ``german''}
	\end{subfigure}%
		\begin{subfigure}{0.5\columnwidth}
		\includegraphics[width=0.98\textwidth]{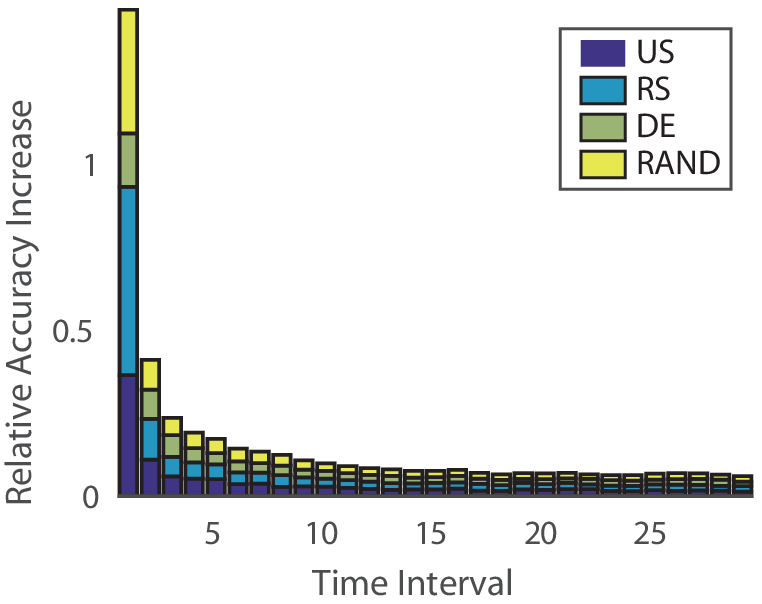}
		\caption{Relative Accuracy\newline Increments: ``german''}
	\end{subfigure}
    \caption{Examples of non-stationary AL in UCI datasets ``fourclass'' and ``ILPD'' using four algorithms/criteria: US, RS, DE, and RAND. Above: Proportion of times each criterion generates the largest increase in accuracy. Below: Relative increase in accuracy. In the relative part all increments are re-scaled by subtracting the minimum increment of accuracy over all criteria in each bin.}
    \label{fig:UCI ND}
\end{figure}

The fundamental limitation of existing MAB-based approaches to AL is that their underlying MAB algorithms do not take into account the temporal dynamics of active learning: different criteria are effective at different learning stages \cite{Donmez2007,journals/tkde/HospedalesGX13}. This issue is illustrated by Fig.~\ref{fig:UCI ND}(a,c,e), where the most effective criterion varies across the entire time horizon. On \textit{fourclass}, Density (DE) sampling is slightly better at  first  and uncertainty (US) is consistently good later on. Similarly in \textit{ILPD} or \textit{german}, representative (RS) and density (DE) sampling are better at the crucial early stage before uncertainty becomes better.
A second issue is that the scale of an accuracy-based reward falls dramatically over time (Fig.~\ref{fig:UCI ND}(b,d,f)). Because of this stationary bandit learners will be unduly biased by the high reward from an initial observation and fail to adapt subsequently. {For example in ILDP a stationary learner may fail to make the switch from DE to US as later rewards in favour of US are  small in scale compared to the initial reward in favour of DE.}

Therefore there are non-stationary aspects both in reward scale, and in reward distribution per-arm (MAB perspective) or per-expert (MAB with expert advice perspective). Thus the MAB problem is  formally non-stationary, violating a fundamental assumption required to guarantee existing MAB algorithms' optimality bounds. 

Here we develop a performance guaranteed stochastic MAB with \textcolor{black}{expert advice}\footnote{\textcolor{black}{We use terminology from \cite{auer2002nonstochastic}. It also has other names, including `contextual bandit' \cite{beygelzimer2011contextual,NIPS2007_3178}, `partial-label problem' \cite{Kakade08efficientbandit}, and `associative bandit problem' \cite{Strehl:2006:ELA:1143844.1143956}.}} algorithm in a \emph{non-stationary} environment. 
Applying this to AL means that, like \cite{hsu2015active}, if there is a single best (but a priori unknown) AL algorithm for a dataset, we are able to quickly discover it and thus approach the performance of an oracle that knows the best algorithm for each dataset. But importantly when different algorithms' efficacies vary over time within one dataset, we can adapt to this and approach the performance of an oracle that knows the best AL algorithm \emph{at each iteration}. 


\section{Background and Related Work}

\subsection{Active Learning}
We denote the pool of data with $M$ samples as $\mathcal{D}=\{\bm{x}_i,l_i,\dots,\bm{x}_M,l_M\}$ where the instances are $\bm{x}_i\in\mathds{R}^d$ and the labels are $l_i\in\{1,\dots,C\}$. In an active learning scenario, the data $\mathcal{D}$ are initially a labelled set $\mathcal{L}$ and unlabelled set $\mathcal{U}=\mathcal{D}\setminus\mathcal{L}$ where $|\mathcal{L}|\ll|\mathcal{U}|$. Training an initial classifier $f_0$ on the samples in the initial set $\mathcal{L}$, the algorithm starts to query  instances $\bm{x}_q$ from $\mathcal{U}$ during  iterations $t=1,\dots,T$. After the supervision $l_q$ of instance $\bm{x}_q$ is obtained,  $\bm{x}_q$ is removed from the unlabelled set $\mathcal{U}$ and added to the labelled set $\mathcal{L}$, from which classifier $f_t$ is retrained.

\subsection{Bandit Algorithms}

\vspace{0.2cm}\noindent\textbf{Multi-armed Bandit}\quad 
In multi-armed bandit (MAB) problems, a player pulls a lever from a set $\mathcal{K}=\{1,\dots,K\}$ of slot machines in a sequence of time steps $\mathcal{T}=\{1,\dots,T\}$ to maximise her payoff. During the game, she only observes the reward $r^k(t)\in[0,1]$ of the specific arm pulled $k$ at time step $t$. The aim of {the player} is to maximise their return, which is the sum of the rewards over the sequence of pulls. This requires a trade-off between exploration (collect information to  estimate the arm with the highest return) and exploitation (focus on the arm with the highest estimated return). {Training a bandit learner to solve a MAB problem is then formalized as minimising the \emph{regret} between the actions chosen by the player's strategy $a_k\sim\pi$, and the best arm.}

For example, the EXP3 algorithm \cite{auer2002nonstochastic} minimises, for any finite $T$, the ``\emph{static} regret'' between the player's reward and the best arm in retrospect: $\max_{k} \sum_{t=1}^T r^k(t)  - \mathbb{E}(\sum_{t=1}^T r^{\pi}(t))$.

\vspace{0.2cm}\noindent\textbf{Contextual Multi-armed Bandit}\quad 
The goal of contextual bandits is to build a relationship between available context information $\bm h_t\in\mathds{R}^d$ and the reward distribution over all arms. For example, LinUCB \cite{Chu11contextualbandits} makes the linear realizability assumption that there exists an unknown weight vector $\bm \theta^*\in\mathds{R}^d$ with $||\theta^*||\leq 1$ so that regret  $\sum_{t=1}^Tr_{a_t^*}(t)-\sum_{t=1}^Tr_{a_t}(t)$ is minimized, where $r_{a_t^*}(t)=\bm \theta^{*\top} \bm h_t$ and $r_{a_t}(t)=\bm \theta^\top \bm h_t$.
However, learning to predict the reward for each data point accurately appears to be an even harder problem given the limited information from only expert suggestions (Fig~\ref{fig:UCI ND}). More importantly, given the changing reward distribution over time, there is no constant relation between context and reward.


\vspace{0.2cm}\noindent\textbf{Multi-armed Bandit with Expert Advice}\quad 
\textcolor{black}{Expert information about the likely efficacy of each arm is often available.} \cite{auer2002nonstochastic} thus introduced an adversarial MAB with expert advice algorithm EXP4 that exploits $N$ experts giving advice vectors (probabilities $\bm \xi^{n}(t) \in [0,1]^K$ over levers) to the learner at each time. In contrast to MAB without expert advice, the goal is now to identify the best expert rather than the best arm. In this setting the regret to minimise is the difference between the return of the best expert in retrospect and the player:
\begin{align}
\label{eq:exp4}
 \max_{n} \sum_{t=1}^T y_t^{n}  - \mathbb{E}(\sum_{t=1}^T y_t^{\pi})
\end{align}
\noindent where $y_t^{n} = \sum_{k=1}^K \xi_k^{n}(t)\times r^k(t)$  is the expected reward of an expert and $y_t^\pi$ is the expected reward of our policy. 

\subsection{Bandits for Active Learning}
For active learning using a MAB with expert advice algorithm, the $N$ experts correspond to our ensemble of active learning criteria and the $K$ arms are  available points in the pool. Each expert (criterion) $n$ provides a probability vector encoding preference $\bm \xi^n(t)$ over arms (instances). Active learners based on MAB with expert advice aim to learn the best criterion for a specific dataset. 
In COMB \cite{Baram:2004:OCA:1005332.1005342}, the authors propose to use MAB with expert advice in active learning and heuristically designed the classification entropy maximization (CEM) score as the reward of the EXP4 bandit algorithm \cite{auer2002nonstochastic}. A more recent paper \cite{hsu2015active} (ALBL) proposed to replace the CEM reward with an unbiased estimation of test accuracy Important Weighted Accuracy (IWA) and used an upgraded bandit algorithm EXP4.P \cite{beygelzimer2011contextual}, which improves the earlier EXP4 method. Similarly, another recent paper \cite{DBLP:journals/corr/ChuL16} applied linear upper confidence bound contextual bandit algorithm (LinUCB) to train an ensemble and transferred the knowledge to other datasets. All of these algorithms enable the selection of a suitable active learning criteria for a given dataset. Our contribution is also to perform AL in a dataset-specific way by optimally tuning the exploration and exploitation of an ensemble of AL algorithms; but more importantly to do so dynamically, thus allowing the optimal tuning to vary as learning progresses. Unlike \cite{Baram:2004:OCA:1005332.1005342,hsu2015active,DBLP:journals/corr/ChuL16} we are able to deal with the non-stationary nature of this process. And unlike the heuristics in \cite{Donmez2007,journals/tkde/HospedalesGX13}, we have a theoretical guarantees, and can work with more than two criteria.

\subsection{Non-stationary Property of Active Learning}
\noindent\textbf{Demonstration of Non-stationarity}\quad
We describe a preliminary experiment to demonstrate empirically the existence of non-stationary reward distributions for a MAB formalisation of AL. Following the learning trajectory of our method, we use an oracle to score all the available query points at each iteration (i.e., hypothetically label each point, update the classifier, and check the test accuracy). Using the actual test accuracy as the reward, we can obtain the true expected reward of the $n$th expert $y_{t}^{n}=\bm \xi^n(t) \bm r(t)$ at each time step $t$. Fig.~\ref{fig:UCI ND} summarises the resulting average reward obtained in every 10 iterations of AL. Based on this, we can further compute the proportion of times that each criterion would obtain the highest reward. It can be seen that the MAB problem is non-stationary as the rewards vary systematically, and there is not a single criterion (expert) which obtains the highest proportion of wins throughout learning. Additionally, the ideal combination of criteria varies across datasets. For example, as illustrated in Fig~\ref{fig:UCI ND}, density and uncertainty sampling show better complementary in ILPD, while  representative and uncertainty sampling are more complementary in german dataset.

\vspace{0.2cm}\noindent\textbf{Existing MAB ensembles are not robust to non-stationarity}\quad
The non-stationary property in the MAB formalisation of AL also highlights the key weakness of COMB and ALBL: they use EXP4/EXP4.P \cite{auer2002nonstochastic,beygelzimer2011contextual} expert advice bandit algorithms which provide guarantees against an inappropriate (static) regret that is only relevant in a stationary problem. In a non-stationary problem, it is clear that even an algorithm that perfectly estimates the best single expert (optimal w.r.t static oracle Eq.~\ref{eq:exp4}) can be arbitrarily worse than one which can choose the best expert at each step (optimal w.r.t dynamic oracle).  In this paper, we develop an non-stationary stochastic MAB algorithm REXP4 {(Restarting Exponential-weight algorithm for Exploration and Exploitation using Expert advice)} with bounds against a stricter dynamic oracle notion of optimality more suited for (non-stationary) AL. 


\vspace{0.2cm}\noindent\textbf{Prior attempts at non-stationary active learners}\quad
A few previous active learning studies also observed that different algorithms are effective at different stages of learning and proposed heuristics for switching two base query criteria (e.g., density sampling at an early stage, and uncertainty sampling later on) \cite{Donmez2007,journals/tkde/HospedalesGX13}. But these only adapt 2 criteria (density and uncertainty) unlike MAB ensembles which learn to combine many criteria, and their heuristics do not provide a principled and optimal way to learn when to switch. 

\vspace{0.2cm}\noindent\textbf{Prior attempts at non-stationary MABs}\quad
Some previous studies have extended MAB without expert advice learning to the non-stationary setting \cite{garivier2008upper,besbes2014stochastic} and provided regret bounds to guarantee the algorithms' performance. However bandits with expert advice are preferable because they can achieve tighter learning bounds  \cite{auer2002nonstochastic,hsu2015active} and they do not treat each criterion as a black box, so that one observation can be informative about many arms. Consider an AL situation where two criteria prefer the same instance. In the MAB interpretation (criteria=arms), after observing a reward, you only learn about the criterion/arm chosen at that iteration. In the MAB with expert advice interpretation (criteria=experts), the observed reward generates updates about the efficacy of all criteria that expressed opinions about the point.


Those few MABs extended to the non-stationary setting have other stronger assumptions. For example, the discounted/sliding-window UCB algorithm \cite{garivier2008upper} assumes the nature of the non-stationarity is that the reward distribution is piece-wise and the number of changes is known. Similarly \cite{Yu:2009:PBP:1553374.1553524} makes the easier piecewise assumption, and also that the retrospective rewards for un-pulled arms are available -- but they are not in active learning. 
In \cite{NIPS2016_6536}, the authors proposed to measure the total statistical variance of the consecutive distributions at each time interval. Their result provides a big picture of the regret landscape for full information and bandit settings. Their proposed method addresses non-stationary environments but only for the regular MAB problem. Despite the use of the term expert in the title, it does not address the Expert-advice variant of the MAB problem relevant to us. It addresses arms rather than experts over arms. 

{We propose a non-stationary MAB with expert advice algorithm that has performance guarantees, and validate its practical application to active learning.}



\section{Non-stationary Multi-Armed Bandit with Expert Advice for Active Learning}

\subsection{Non-stationary Multi-Armed Bandit with Expert Advice: REXP4}
To formalise the problem, we assume the expected reward $y_t^n$ of each expert $n$ can change at any time step $t$. The total variation of the expected reward over all $T$ steps is
\begin{align}
    \sum_{t=1}^{T-1} \sup_n |y_t^n-y_{t+1}^n|
\end{align}
Following \cite{DBLP:journals/corr/BesbesGZ13,besbes2014stochastic}, we assume this total variation in expected reward is bounded by a variation budget
$V_T$. 
The variation budget captures our assumed constraints on the non-stationary environment. It allows a wide variety of reward changes -- from continuous drift to discrete jumps -- yet provides sufficient constraint to permit a bandit algorithm to 
learn in a non-stationary environment.
Temporal uncertainty set $\mathcal{V}$  is defined as the set of reward vector
sequences that are subject to the variation budget $V_T$ over all $T$ steps.
\begin{equation*}
\mathcal{V}=\Bigg\{ y\in [0,1]^{N\times K}: \sum_{t=1}^{T-1} \sup_{n} |y_t^n - y_{t+1}^n | \leq V_T  \Bigg\}
\end{equation*}


To bound the performance of a bandit learner in a non-stationary environment, we work with the regret between the learner
 and a \emph{dynamic} oracle. The regret is defined as the worst-case difference between the expected policy return and the return of using the best expert at each time $t$.

\begin{defn}
Dynamic Regret for Multi-Armed Bandit with Expert Advice
\begin{equation}
R^{\pi} (\mathcal{V}, T)  = \sup_{y\in \mathcal{V}} \{\sum_{t=1}^T y_t^{*} - \mathbb{E}^{\pi} [\sum_{t=1}^T y_t^{\pi}]\}\label{eq:dynamicRegret}
\end{equation}
\end{defn}
\noindent where $y_t^{*} = \max_{n} y_t^{n}$ is the best possible expected reward among all experts at time $t$. Our regret is against this dynamic oracle, in contrast to prior MABs' static oracle (Eq~\ref{eq:exp4}).


Our non-stationary MAB with expert advice algorithm REXP4 minimises the dynamic regret in Eq~\ref{eq:dynamicRegret}. As shown in Algorithm~\ref{Rexp4}, it trades off between the need to remember and forget by breaking the task into batches and applying EXP4 \cite{auer2002nonstochastic} on each batch. As the reward distribution changes, it adapts to the change as by re-estimating each expert's reward distribution at each batch. We show the worst case bound on the regret between this REXP4 procedure and the dynamic oracle.

\subsection{Regret Bound for REXP4}
The regret bound for REXP4 
is illustrated in the following theorem. The theorem is proved by following the proof structure of \cite{besbes2014stochastic} and replacing the term $\mu$ in \cite{besbes2014stochastic} with the expected reward term $y$ in our paper.

\begin{thm}
Let $\pi$ be the REXP4 policy with a batch size $\Delta_T = \lceil (A \log N)^{1/3} (T/V_T)^{2/3} \rceil$ and $\gamma = \min \{1, \sqrt{\frac{A\log N}{(e-1)\Delta_T}}\}$. Then, there is some constant $C$ such that for every $T \geq 1, K\geq 2, N \geq 2$, and $V_T\in[A^{-1},A^{-1}T]$
\begin{equation}
R^{\pi}(\mathcal{V},T) \leq C (A \log N \cdot V_T)^{1/3} T^{2/3}
\label{eq:bound}
\end{equation}
where $A=\min \{N, K\}$ indicates the smaller number of experts or arms.
\end{thm}

The result is an upper bound on the regret between our REXP4 policy and the dynamic oracle. As $A=\min\{N,K\}$, it is favourable if either the number of experts $N$ or arms $K$ is small. This also means it is relatively robust to many arms (as in AL, where arms=data points). If $V_T$ is sub-linear in $T$ (total variation in reward grows slower than timesteps), then performance converges to that of the oracle.

\begin{algorithm}[t]
	\caption{Pseudocode of algorithm REXP4}
	\label{Rexp4}
	\begin{algorithmic}
		\Statex \textbf{Inputs:} $\gamma\in (0,1]$ and an epoch size $\Delta_T$
		\begin{enumerate}
			\item Set Epoch index $j=1$
			\item Repeat while  $j \leq\lceil T/\Delta_T \rceil$
			\begin{itemize}
				\item Set $\tau = (j-1) \Delta_T$
				\item Initialisation: for any expert $n$ set weight $w_n(t) = 1$
				\item Repeat for $t = \tau+1, \dots, \min \{T, \tau + \Delta_T\}$, Call EXP4 Algorithm\cite{auer2002nonstochastic}
				\item Set $j= j+1$ and return to the beginning of step 2
			\end{itemize}
		\end{enumerate}
	\end{algorithmic}
\end{algorithm}

\begin{algorithm}[t]
\caption{DEAL: Dynamic Ensemble Active Learning}
\label{AL with Rexp4}
\begin{algorithmic}
\Statex \textbf{Inputs:} $\gamma\in (0,1]$, initial weight $\bm w(1) =1$, $\Delta_T=10$, $\tau=1$,labelled set $\mathcal{L}_0$, unlabelled set $\mathcal{U}_0$, initial classifier $f_0$
\For{$t=1\rightarrow T$}
\begin{enumerate}
            \item Get scores of instance $\bm s^{n}_t$ from criteria
            \item Normalised the score vector $\bm s_t^n=-\exp(-\alpha \text{ } \bm {rank})$
            \item Obtain the advise vector with $\bm \xi^n(t)=\frac{\exp(-\beta\bm s_t^n)}{\sum_k \exp(-\beta s_{t,k}^n)}$
            \item Set $W_t= \sum_{n=1}^{N} w_n(t)$ and for $k=1,\dots, K$ set
            \begin{equation*}
            p_k(t) = (1-\gamma) \sum_{i=n}^N \frac{w_n(t) \xi_k^n(t)}{W_t} +\frac{\gamma}{K}
            \end{equation*}
            \item Query the label of instance $\bm x_{k_t}$ randomly from $\mathcal{U}_t$ according to  probability $p_1(t),\dots, p_K(t)$
            \item Move the instance $\bm x_{k_t}$ from $\mathcal{U}_t$ to $\mathcal{L}_t$
            \item Retrain the classifier $f_t$ and receive reward $r_t^k \in [0,1]$
            \item For $k=1, \dots, K$ set
            \begin{equation*}
                \hat r_k(t) = \left\{ \begin{array}{ll}
                r_k(t)/p_k(t) & \textrm{if } k = k_t\\
                0 & \textrm{otherwise}
               \end{array} \right.
            \end{equation*}
            \item For $n =1,\dots, N$ set
            \begin{equation*}
                \begin{split}
                 & \hat y_t^n = \bm \xi^{n}(t)^T \hat{\bm r}(t)\\
                 & w_n(t+1) = w_n(t) \exp (\gamma \hat y_t^n/ K)
                \end{split}
            \end{equation*}
            \item $\tau=\tau+1$

\end{enumerate}
            \If{$\tau>\Delta_T$}
            Reset $\tau=1$ and $\bm w(t+1)=\bm 1$
            \EndIf
\EndFor
\end{algorithmic}
\end{algorithm}
\subsection{Dynamic Ensemble Active Learning}

\begin{figure}
    \centering
    \includegraphics[width=1.0\columnwidth]{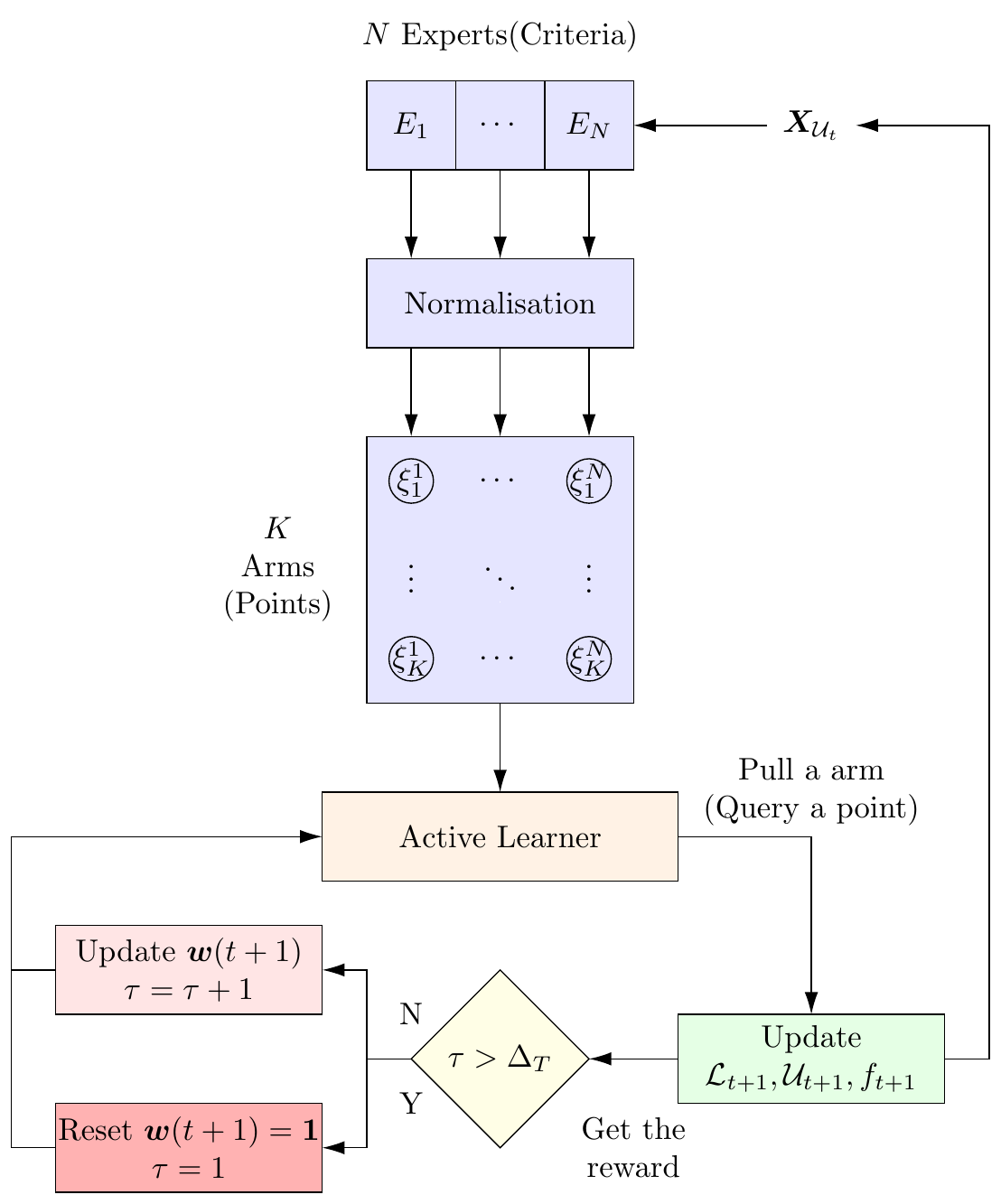}
    \caption{Illustration of DEAL System. Light blue: Taking the unlabelled set $\bm X_{\mathcal{U}_t}$ as the input,  each expert will output a score that is normalised before input to the DEAL active learner. $\xi_K^N$ is the $N$th criterion score of $K$th instance. Orange: the active learner to make a decision. Green: updating the labelled set, unlabelled set, and the classifier. Light yellow: The restart detection scheme. Ensemble weights are then updated differently between (light red) or at (dark red) restarts. }
    \label{fig:DEALsys}
\end{figure}

Based on our REXP4 algorithm for MAB with expert advice, we present DEAL-REXP4 (Dynamic Ensemble Active Learning) for active learning based on REXP4. Our dynamic ensemble learner will update both base learner $f_t$ and active criteria weights $\bm w(t)$ iteratively. More specifically,  each ensemble criterion will predict scores $\bm s_{t}^n$ for all unlabelled instances. We use exponential ranking normalisation $-\exp(-\alpha \text{ } rank)$ to avoid the issue of different criterion scales, and apply the Gibbs measure $\frac{\exp(-\beta\bm s_t^n)}{\sum_k \exp(-\beta s_{t,k}^n)}$ where the parameters $\alpha,\beta$ control the sharpness of the distribution. The $rank$ denotes the ranking position of the instance's score where the ranking order is determined by the criterion strategy's ordering. For example, the entropy criterion prefers points with maximum entropy, so the maximum entropy point has rank 1. Similarly, the minimum margin criterion prefers points with low distance to margin, so the minimum distance point has rank 1.
Based on the current suggestions from the criteria members, the active learning ensemble $\bm w(t)$ will select an instance for label querying. Then, the base learner $f_{t+1}$ will be updated with the new labelled data and the active learner $\bm w(t+1)$ will be updated successively based on the performance improvement of the updated base learner. To learn the non-stationary reward distribution, we use our proposed REXP4 algorithm to learn the weights of active learning criteria in an online adaptive way by introducing the restart scheme. Giving the current within-batch index $\tau\in\{1,\cdots,\Delta_T\}$, the restart scheme will be activated when $\tau>\Delta_T$, otherwise updates follow the EXP4 rule. The details are described in Algorithm~\ref{AL with Rexp4} with an illustration in Fig. \ref{fig:DEALsys}.


In DEAL-REXP4 we set the reward as the resulting accuracy after a classifier update. Thus in the context of active learning, the bound given in Eq.~\ref{eq:bound} means that we know that the total area under the reward curve obtained by DEAL-REXP4 is within a bound of the best case scenario that would occur only \emph{if we had known the best criterion to use at each iteration.} Moreover, if the variation budget $V_T$ grows sub-linearly with $T$, DEAL-REXP4 converges towards this best-expert-per-iteration upper bound scenario.

\begin{table*}[t]
	\centering
	\caption{Summary of Active Learning Algorithms}
	\resizebox{\textwidth}{!}{
	\begin{tabular}{llccccc}
		\hline
		\multicolumn{5}{l}{\textbf{Single Criterion}} \\
		Algorithm & \multicolumn{2}{c}{Motivation} & Stationarity & Importance of Criterion & Ensemble Members & Property\\
		\hline
		US \cite{Lewis:1994:SAT:188490.188495,settles2009active, conf/cvpr/2009}  & \multicolumn{2}{l}{Querying the least confidence} & Stationary & Fixed & US & Static\\
		RS \cite{Xu:2003:RST:1757788.1757826} & \multicolumn{2}{l}{Query a cluster within Margin} & Stationary & Fixed & RS & Static\\
		DE \cite{Donmez2007} & \multicolumn{2}{l}{Query the major cluster} & Stationary & Fixed & DE & Static\\
		\hline
		\multicolumn{5}{l}{\textbf{Multiple Criteria}} \\
		Algorithms & \multicolumn{2}{c}{Motivations} & Stationarity & Importance of Criterion & Ensemble Members & Property\\
		\hline
		
		QUIRE \cite{conf/nips/2010} & \multicolumn{2}{l}{Combining informativeness and representativeness} & Stationary & Equal effect & QUIRE & Static\\
		BMDR \cite{Wang:2015:QDR:2737800.2700408} & \multicolumn{2}{l}{Combining discriminative and representativeness} & Stationary & Equal effect & BMDR & Static\\
		LAL \cite{DBLP:journals/corr/KonyushkovaSF17} & \multicolumn{2}{l}{Combining Multiple motivations} & Stationary & Equal effect & Any Criteria & Static\\
		DUAL\cite{Donmez2007} & \multicolumn{2}{l}{Switching from DE to US once} & Non-stationary & Varying & US, DE & Dynamic\\
		ALGD \cite{journals/tkde/HospedalesGX13} & \multicolumn{2}{l}{Switching between DE to US} & Non-stationary & Varying &US, DE & Dynamic\\
		\hline
		\multicolumn{5}{l}{\textbf{Bandit Ensemble Algorithms}} \\
		Algorithm & Bandit & Regret & Stationarity & Importance of Criterion & Ensemble Members & Property \\
		\hline
		COMB \cite{Baram:2004:OCA:1005332.1005342} & EXP4 \cite{auer2002nonstochastic} & $\operatorname{max}_n\sum_{t=1}^Ty_t^n-\mathds{E}(\sum_{t=1}^Ty_t^\pi)$ & Stationary & Single best & Any Criteria & Static \\
		ALBL \cite{hsu2015active} & EXP4.P \cite{beygelzimer2011contextual} & $\operatorname{max}_n\sum_{t=1}^Ty_t^n-\mathds{E}(\sum_{t=1}^Ty_t^\pi)$ & Stationary & Single best & Any Criteria & Static\\
		LSA \cite{DBLP:journals/corr/ChuL16} & LinUCB \cite{Chu11contextualbandits} & $\sum_{t=1}^Tr_{a_t^*}(t)-\sum_{t=1}^Tr_{a_t}(t)$ & Stationary & Single best combination & Any Criteria & Static\\
		DEAL & REXP4 & \textcolor{black}{$\sum_{t=1}^T\operatorname{max}_ny_t^n-\mathds{E}(\sum_{t=1}^Ty_t^\pi)$} & Non-Stationary & \textcolor{black}{Dynamic best} & Any Criteria & Dynamic\\
		\hline
	\end{tabular}}
	\label{tab:sumal}
\end{table*}

\subsection{Discussion of Static and Dynamic Active Learning}

We divide active learning algorithms into static/dynamic based on the stationary/non-stationary assumption on the importance of each criteria over different time periods.


\vspace{0.2cm}\noindent\textbf{Static Active Learning}\quad Single criterion algorithms are all static, since they solve active learning with only one criterion. Regarding active learning algorithms with multiple motivations: if they are formalised as a single fixed mixture of criteria, they are also static. Since the coefficients of different motivations are fixed over all time steps, they assume that a single weighted combination is suitable at any learning stage.
For example, Query Informative and Representative Examples (QUIRE) \cite{conf/nips/2010}, Learning Active Learning (LAL) \cite{DBLP:journals/corr/KonyushkovaSF17}, and Discriminative and Representative Queries for Batch Mode Active Learning (BMDR) \cite{Wang:2015:QDR:2737800.2700408} are static active algorithms with multiple motivations.


Previously proposed ensemble algorithms  ALBL \cite{hsu2015active}, COMB \cite{Baram:2004:OCA:1005332.1005342}, and Linear Strategy Aggregation (LSA) \cite{DBLP:journals/corr/ChuL16} are also static in the sense that, although the weight proportion of their ensemble members changes as data is gathered, their underlying bandit learner is a stationary one, assuming there is only one best expert or best linear combination over all time. 


\vspace{0.2cm}\noindent\textbf{Dynamic Active Learning}\quad In our dynamic active learning research question, we avoid a stationarity assumption on criteria importance over time. A non-stationary algorithm should adapt its weighting proportions over time in response to  learning progress. Prior attempts propose heuristics for classifier switching or reweighting  \cite{Donmez2007,journals/tkde/HospedalesGX13} between density and uncertainty sampling. Our DEAL-REXP4 improves on these in that it can use an arbitrary number of criteria of any type beyond 2 specified criteria; and in contrast to prior heuristics, it contains a principled underlying learner with theoretical guarantees.
We provide a summary of related prior active learning algorithms  in Table~\ref{tab:sumal}, where the generality and strong notion of regret in DEAL-REXP4 is clear.


\begin{figure}[t]
	\centering
	\begin{subfigure}{0.49\columnwidth}
		\centering
		\includegraphics[width=\linewidth]{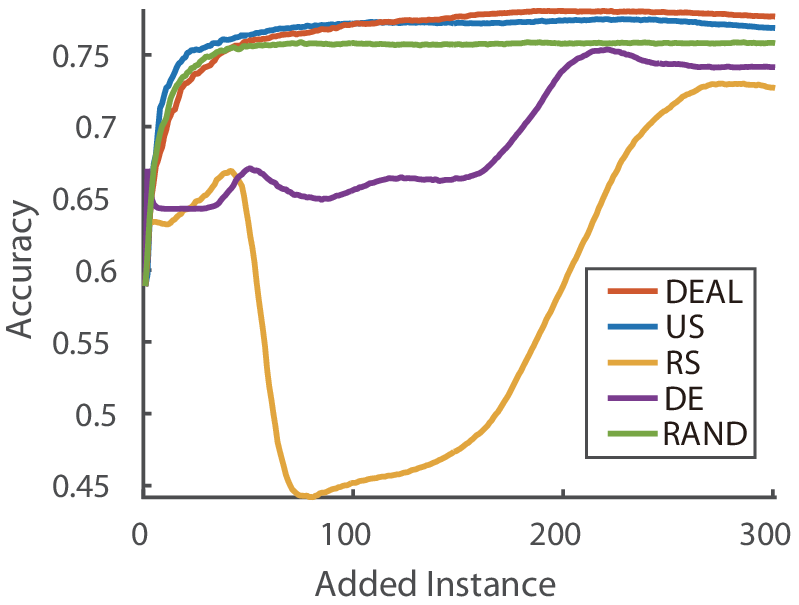}
		\caption{fourclass}
	\end{subfigure}
	\begin{subfigure}{0.49\columnwidth}
		\centering
		\includegraphics[width=\linewidth]{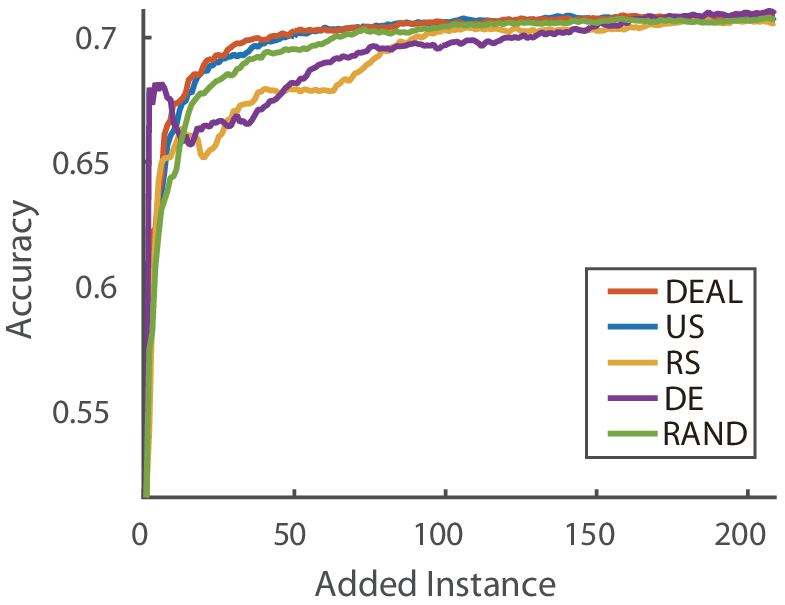}
		\caption{ILPD}
	\end{subfigure}
	\caption{Comparison of DEAL-REXP4 versus individual ensemble members.}
	\label{fig:bin ns criteria comparison}
\end{figure}

\begin{figure}[t]
	\centering
	\begin{subfigure}{0.49\columnwidth}
		\centering
		\includegraphics[width=1\linewidth]{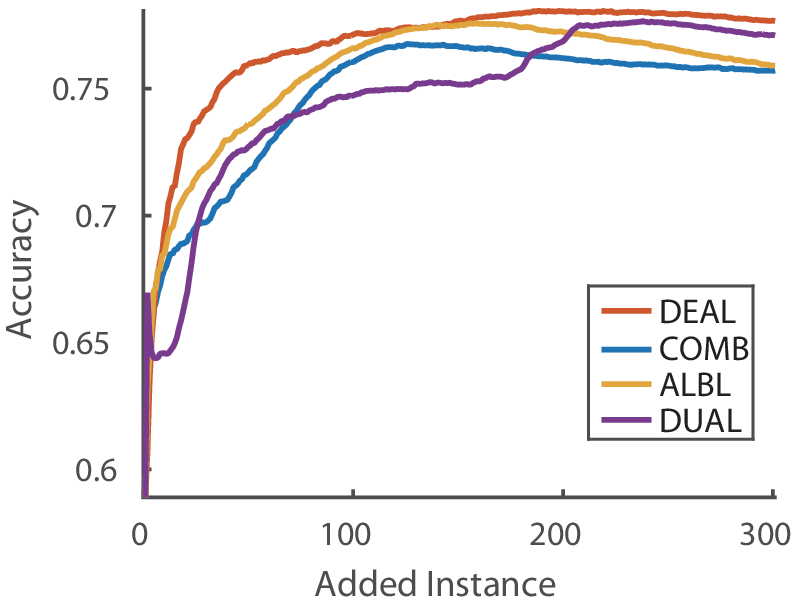}
		\caption{fourclass}
	\end{subfigure}%
	\begin{subfigure}{0.49\columnwidth}
		\centering
		\includegraphics[width=1\linewidth]{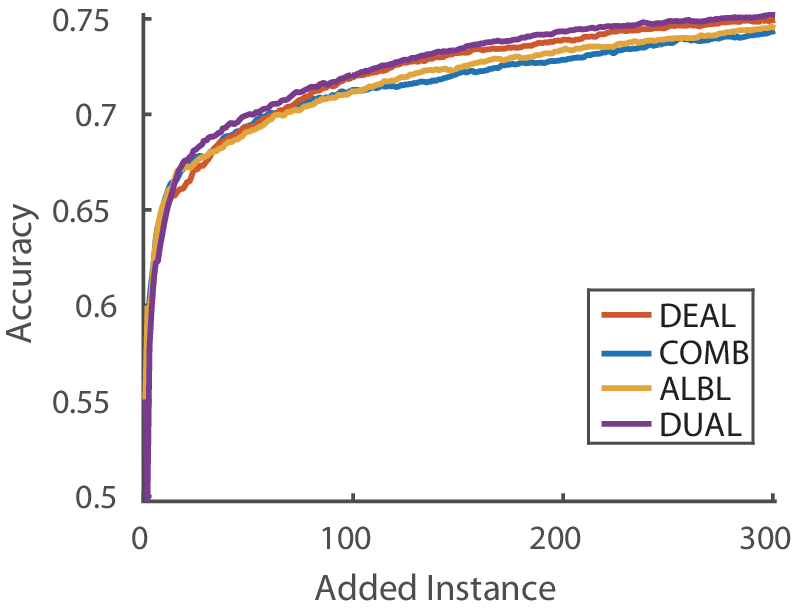}
		\caption{german}
	\end{subfigure}
	\begin{subfigure}{0.49\columnwidth}
		\centering
		\includegraphics[width=1\linewidth]{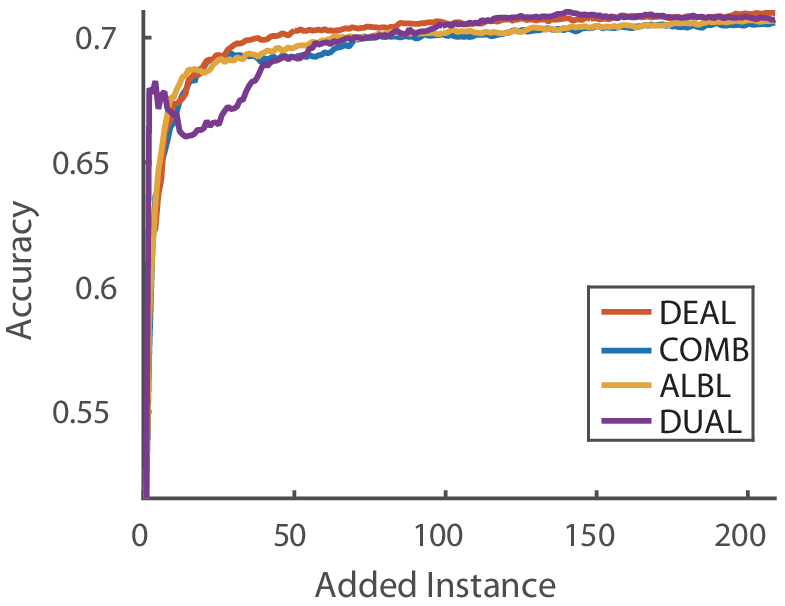}
		\caption{ILPD}
	\end{subfigure}
	\begin{subfigure}{0.49\columnwidth}
		\centering
		\includegraphics[width=1\linewidth]{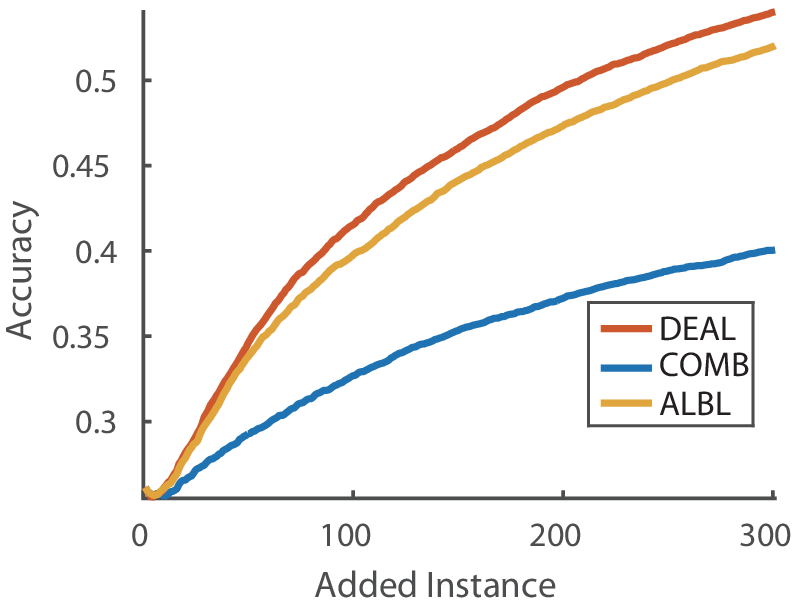}
		\caption{letter}
	\end{subfigure}
	\caption{Comparison of active learning with our DEAL-REXP versus alternative state of the art bandit algorithms.}
	\label{fig:bin bandit comparison}
\end{figure}

\section{Experiments and Results}

\begin{table*}[htbp]
\centering
\caption{Win/Tie/Loss counts of DEAL-REXP4 versus ensemble members in terms of AUC at specified learning stage.}
\resizebox{1.0\textwidth}{!}{
\begin{tabular}{ccccccccccccc}
\hline
Rank & $5\%$ & $10\%$& $15\%$& $20\%$ & $25\%$ & $30\%$ & $35\%$ & $40\%$ & $45\%$ & $50\%$ & Total\\
\hline
1st & $0/4/9$  & $0/3/10$ & $0/3/10$ & $0/2/11$ & $0/2/11$ & $0/4/9$  & $0/4/10$ & $0/3/10$ & $0/3/10$ & $0/3/10$  & $0/30/100$\\
2nd & $2/6/5$  & $4/5/4$  & $4/6/3$ & $5/4/4$  & $6/3/4$ & $6/3/4$  & $6/3/3$ & $6/4/3$  & $6/4/3$ & $5/6/2$  & $50/45/35$  \\
3rd & $7/5/1$  & $7/4/2$  & $7/5/1$ & $7/5/1$  & $7/4/2$ & $7/4/2$  & $7/6/0$ & $7/6/0$  & $8/5/0$ & $8/5/0$  & $72/49/9$  \\
4th & $11/2/0$ & $12/1/0$ & $13/0/0$ & $13/0/0$ & $13/0/0$ & $13/0/0$ & $13/0/0$ & $13/0/0$ & $13/0/0$ & $13/0/0$ & $127/3/0$  \\
\hline
Total & $20/17/15$ & $23/13/16$ & $24/14/14$ & $25/11/16$ & $26/9/17$ & $26/11/15$ & $26/13/13$ & $26/13/13$ & $27/12/13$ & $26/14/12$ & $249/127/144$\\
\hline
\end{tabular}}
\label{tab: bin criteria t test}
\end{table*}

\begin{table*}[htbp]
\centering
\caption{Win/Tie/Loss counts of DEAL-REXP4 and state of the art alternatives at specified learning stages.}
\resizebox{1.0\textwidth}{!}{
\begin{tabular}{cccccccccccc}
\hline
Algorithm & $5\%$ & $10\%$ & $15\%$ & $20\%$ & $25\%$& $30\%$ & $35\%$ & $40\%$ & $45\%$ & $50\%$ & Total\\
\hline
\multicolumn{12}{l}{\textbf{Non-Stationary Datasets}} \\
\hline
ALBL          & $8/15/3$ & $9/13/4$ & $9/12/5$  & $9/11/6$ & $7/13/6$ & $6/14/6$ & $6/15/5$ & $4/16/6$  & $4/16/6$  & $4/16/6$  & $66/141/53$  \\
COMB          & $4/13/9$ & $2/14/10$& $2/13/11$ & $2/12/12$ & $1/12/13$ & $2/12/12$ & $2/13/11$& $1/15/10$  & $0/16/10$  & $0/16/10$  & $16/136/108$ \\
DUAL          & $7/9/8$  & $7/7/10$ & $7/7/10$  & $8/6/10$ & $9/8/7$ & $8/8/8$ & $6/11/7$ & $7/12/5$    & $7/14/3$  & $7/15/2$  & $73/97/70$  \\
\textbf{DEAL} & $6/15/5$ & $9/14/3$ & $11/12/3$ & $12/11/3$ & $12/11/3$ & $12/12/2$ & $11/13/2$ & $12/11/3$  & $11/12/3$  & $10/13/3$  & $\bm{106/124/30}$   \\
\hline
\multicolumn{12}{l}{\textbf{Stationary Datasets}} \\
\hline
\textbf{ALBL} & $4/6/1$  & $6/3/2$ & $7/3/1$ & $6/3/2$ & $7/2/3$ & $6/3/2$ & $6/3/2$  & $5/5/1$  & $5/5/1$  & $4/6/1$  & $\bm{56/39/15}$ \\
COMB          & $2/4/5$  & $2/2/7$ & $1/4/6$ & $1/2/8$ & $1/2/8$ & $1/2/8$  & $1/3/7$  & $1/4/6$  & $1/4/6$  & $1/4/6$  & $12/31/67$  \\
DUAL          & $0/2/7$  & $3/1/5$ & $3/3/3$ & $5/2/2$ & $5/2/2$ & $5/2/2$  & $4/4/1$  & $4/4/1$  & $4/4/1$  & $4/4/1$  & $37/28/25$  \\
DEAL          & $7/4/0$  & $6/2/3$ & $3/4/4$ & $4/3/4$ & $4/2/5$ & $4/3/4$  & $3/4/4$  & $2/5/4$  & $2/5/4$  & $2/6/3$  & $37/38/35$   \\
\hline
\end{tabular}}
\label{tab: bin bandit t test}
\end{table*}

To evaluate our algorithm, we use 13 datasets from UCI\footnote{https://archive.ics.uci.edu/ml/datasets.html} and LibSVM\footnote{https://www.csie.ntu.edu.tw/~cjlin/libsvmtools/datasets/binary.html} repositories. \textcolor{black}{These datasets are selected following previous relevant papers \cite{DBLP:journals/corr/ChuL16,hsu2015active,conf/nips/2010,conf/kdd/2012}.} We use linear SVM \cite{Fan:2008:LLL:1390681.1442794} as the base learner. If the datasets do not include a pre-defined training/testing split, we randomly split $60\%$ for training and the rest for testing. In each trial, we start with 1 randomly labelled point per class. Each experiment is repeated 200 times and the average testing accuracy is reported. 

 \vspace{0.2cm}\noindent\textbf{Criteria Ensemble}:\quad The ensemble of base learners includes: \emph{US}: picking the instances with max-entropy (min margin) instance in binary class datasets \cite{Lewis:1994:SAT:188490.188495,settles2009active} or minimum Best-versus-Second-Best (BvSB) \cite{conf/cvpr/2009} in multiclass datasets. \emph{RS}: clustering the points near the margin \cite{Xu:2003:RST:1757788.1757826} then scoring unlabelled points by their distances to the largest centroid. \emph{Distance-Furthest-First (DFF)}: Focuses on exploration by selecting the furthest unlabeled instance to the nearest labeled instance \cite{Hochbaum:1985:BPH:2775965.2775967}. We use \emph{DFF} which selects the furthest unlabelled instance to the nearest labelled instance \cite{Hochbaum:1985:BPH:2775965.2775967} to replace the RS in multiclass datasets as originally RS is designed for binary class datasets. Both are motivated by exploring the datasets, but DFF does not depend on binary classifiers.  \emph{Density Estimation (DE):} Picking the instance with maximum density in a GMM with 20 diagonal covariance components \cite{Donmez2007}. \emph{RAND:} Randomly selecting points can be hard to beat on datasets unsuited to a given criterion. Moreover, including a random expert (for exploration) is necessary to guarantee the performance of the EXP4 subroutine \cite{auer2002nonstochastic,beygelzimer2011contextual}.


\vspace{0.2cm}\noindent\textbf{Competitors}:\quad  We compare our method to ALBL \cite{hsu2015active}, COMB \cite{Baram:2004:OCA:1005332.1005342} and DUAL \cite{Donmez2007}. For COMB, we follow their recommended settings with CEM reward and $\beta=100$. For the ALBL, we use their settings and importance-weighted accuracy reward.

For direct comparison, ALBL, COMB and REXP4  use the same ensemble of criteria described above. DUAL is engineered for a specific pair of criteria, so we apply its original version using Uncertainty Sampling and Density-Weighted Uncertainty Sampling. It is also only defined for binary classification problems unlike the others.

\vspace{0.2cm}\noindent\textbf{DEAL-REXP4 Settings}:\quad For reward, we follow  \cite{hsu2015active,DBLP:journals/corr/ChuL16} in using the IWA for unbiased estimation of test accuracy. To produce probabilistic preferences for points from all AL criteria, we use exponential ranking normalisation and a Gibbs measure with  \textcolor{black}{$\alpha=0.1, \beta=100$}. We use batch size $\Delta_T=10$ throughout. \textcolor{black}{The choice $\Delta_T=10$ is based on observing the typical coarse duration of performance gaps among different criteria. For example, 
RS wins first 20 iterations in Fig.~\ref{fig:bin ns criteria comparison}(b).} 
The reason for parameterizing in terms of $\Delta_T$ rather than $V_T$ is that it has intuitive meaning in AL context (batch-size), yet implies a corresponding variation budget for any given $T$ (Theorem~1).

\vspace{0.2cm}\noindent \textbf{Characterising dataset (non)stationarity:}\quad
We first investigate each dataset to characterise its (non)stationarity. We use our DEAL trajectory, and use an oracle to measure the $\%$ wins of each criterion at each batch $\Delta_T$ in terms of performance increase. A dataset with stationary reward distribution would tend to have a consistent winner, and vice-versa. Although (non)stationarity is a continuum, we will describe a dataset as stationary if at least two criteria have a fraction of wins above  threshold $\theta=10\%$. 

\vspace{0.2cm}\noindent\textbf{DEAL versus Individual Criteria}\quad
Examples comparing the performance of DEAL and individual criteria in the ensemble are shown in Fig.~\ref{fig:bin ns criteria comparison}.  There is no single criterion that works best for all datasets, moreover different criteria are effective at different stages of learning.  While DEAL is not  best across all datasets and all time-steps (this would require the actual dynamic oracle upper bound), it performs well overall. This is summarised quantitatively across all 13 datasets in Tab.~\ref{tab: bin criteria t test}. Each method's performance is evaluated by the area under the learning curve at different proportions of added instances. The results show the number of wins/ties/losses of DEAL versus the alternative ensemble member of specified highest rank according to two-sided t-test. This shows for example that DEAL often ties with the top-ranked ensemble member (30 draws vs 1st rank), is usually at least as good as the second ranked member (50 wins and 45 ties vs only 35 losses) and is never the worst (0 losses vs 4th rank).

\vspace{0.2cm}\noindent\textbf{Comparison vs State-of-the-Art}\quad
We compare our DEAL-REXP4 with state-of-the-art alternatives to tuning an AL-ensemble. Sometimes DUAL performs well, but it is highly variable depending on whether the criterion switch heuristic makes a good choice or not, as seen in Fig.~\ref{fig:bin bandit comparison}. 
Tab.~\ref{tab: bin bandit t test} summarises the results across all datasets in terms of AUC wins/draws/losses of each approach against the alternatives. DUAL has a lower row-total as it is defined for binary problems only, so not evaluated on \textcolor{black}{wine and} letter  datasets.
The main observation is that DEAL outperforms the alternatives particularly on non-stationary datasets. On stationary datasets we are only slightly worse than ALBL. This is expected as REXP4 performs forgetting in order to adapt to changes in expert efficacy, meaning that we cannot exploit the best criterion as aggressively as ALBL's EXP4.P MAB learner. Nevertheless, overall DEAL is fairly robust to stationary datasets (small margin behind ALBL), while ALBL is not robust to non-stationary datasets (larger margin behind DEAL).

\section{Conclusion}
We proposed a non-stationary multi-armed bandit with expert advice algorithm REXP4, and demonstrated its application to online learning of a criterion ensemble in active learning. The theoretical results provide bounds on REXP4's optimality. The empirical results show that active learning with DEAL-REXP4 tends to perform near the best criterion in the ensemble. It performs comparable to state of the art alternative ensembles on stationary datasets, and outperforms them on non-stationary datasets. 

\bibliographystyle{IEEEtran}


\end{document}